\documentclass{article}
\usepackage{ijcai23}

\usepackage{times}
\usepackage{soul}
\usepackage{url}
\usepackage[hidelinks]{hyperref}
\usepackage[utf8]{inputenc}
\usepackage[small]{caption}
\usepackage{graphicx}
\usepackage{amsmath}
\usepackage{amsthm}
\usepackage{booktabs}
\usepackage{algorithm}
\usepackage{algorithmic}
\usepackage[switch]{lineno}

\usepackage{xcolor}
\usepackage{comment}
\usepackage{color}
\usepackage{multirow}
\usepackage{enumitem}
\usepackage{amssymb}
\usepackage{makecell}   

\newcommand{\etal}{{\emph{et al. }}}

\title{NaturalFinger: Generating Natural Fingerprint with Generative Adversarial Networks}

\author{
Kang Yang\and 
Kunhao Lai$^{1}$ \and
\affiliations
$^1$School of Cyber Science and Engineering, Wuhan University, China \\
}

\begin{document}

\maketitle
\begin{abstract}
	
	Deep neural network (DNN) models have become a critical asset of the model owner as training them requires a large amount of resource (i.e. labeled data). Therefore, many fingerprinting schemes have been proposed to safeguard the intellectual property (IP) of the model owner against model extraction and illegal redistribution. However, previous schemes adopt unnatural images as the fingerprint, such as adversarial examples and noisy images, which can be easily perceived and rejected by the adversary. In this paper, we propose NaturalFinger which generates natural fingerprint with generative adversarial networks (GANs). Besides, our proposed NaturalFinger fingerprints the decision difference areas rather than the decision boundary, which is more robust. The application of GAN not only allows us to generate more imperceptible samples, but also enables us to generate unrestricted samples to explore the decision boundary. 
	To demonstrate the effectiveness of our fingerprint approach, we evaluate our approach against four model modification attacks including adversarial training and two model extraction attacks. Experiments show that our approach achieves 0.91 ARUC value on the FingerBench dataset (154 models), exceeding the optimal baseline (MetaV) over 17\%. Our code is available at \url{https://anonymous.4open.science/r/NaturalFinger-E5E1}.

\end{abstract}

\section{Introduction}

In the past few years, deep neural networks (DDNs) have been applied to a wide range of fields like autonomous driving~\cite{auto_driving}, face recognition~\cite{FaceRecognition}, and intelligent healthcare~\cite{healthcare} due to their outstanding performance. While DNNs are prevalent in our lives, training such a model is a non-trivial task as it requires a large amount of annotated data and powerful computing resources. Thus, many companies provide the prediction of their trained model as APIs to profit, such as Machine-Learning-as-a-Service (MLaaS). In other words, the model is becoming a critical business asset, which requires to be protected against illegal redistribution and model extraction. 


Generally, a type of common approach to protect the intellectual property (IP) of DNN models is model fingerprinting~\cite{ipguard,metafinger,conferrableAdv}. This kind of scheme first constructs a query set based on the source model (protected model) as the fingerprint. Then, the defender queries the suspect model with the query set. If the matching rate between the predicted labels of the suspected model and the query set labels exceeds a threshold, the defender judges the suspect model as a stolen model. 

\begin{figure}[t] 
	\centering  

	\includegraphics[width=0.6\linewidth]{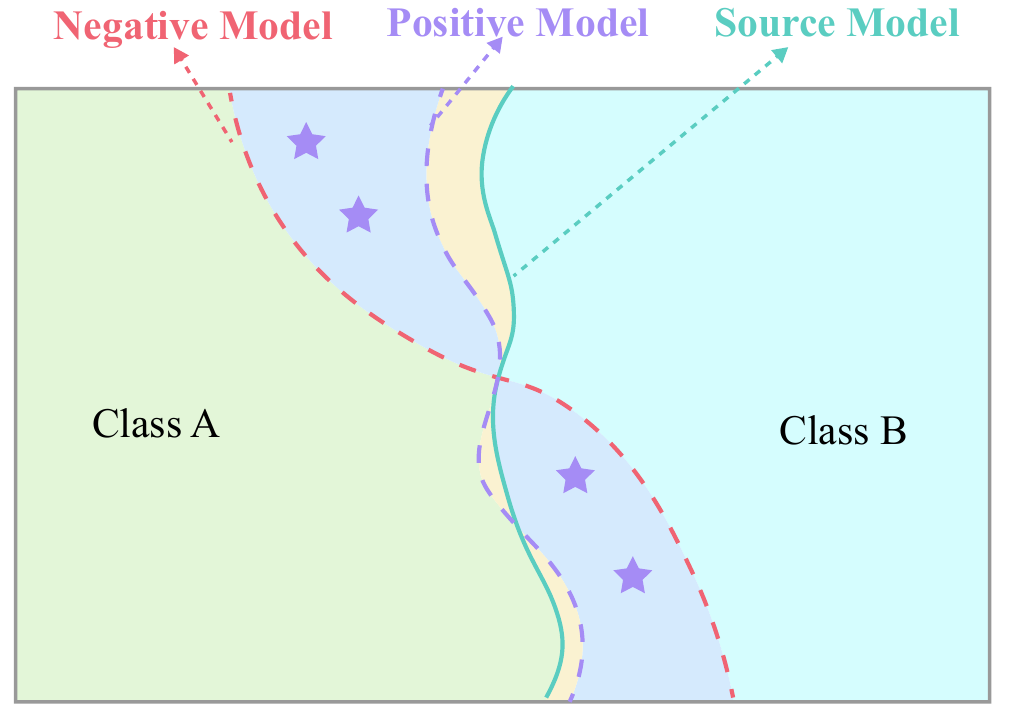}
	
	\caption{The intuition of our proposed NaturalFinger. It leverages GAN to fingerprint the decision difference areas (purple areas) between positive models and negative models.}
	\label{fig:intuition}
\end{figure}

However, previous fingerprinting schemes suffer from the following two problems: 1) Stealthiness. They generate unnatural samples to query the suspect model like noisy examples, which can be perceived and rejected by the adversary; 2) Robustness. Many fingerprinting schemes~\cite{DFfinger,UAPFinger} leverage various adversarial examples to fingerprint the decision boundary, which is not robust against adversarial defense. 

\begin{figure}[t]
	\centering  

	\setlength\tabcolsep{1pt}
	\scriptsize
	\begin{tabular}{cccc}
		\includegraphics[width=0.24\linewidth]{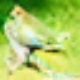} &
		\includegraphics[width=0.24\linewidth]{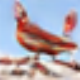} &
		\includegraphics[width=0.24\linewidth]{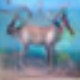}&
		\includegraphics[width=0.24\linewidth]{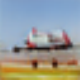} \cr

		(a) Frog/Bird & (b) Bird/Airplane & (c) Horse/Deer & (d) Airplane/Ship
	\end{tabular}
	\caption{The examples of NaturalFinger. The first label and the second label are predicted by positive models and negative models, respectively.}
	
	\label{fig:demo}
\end{figure}

To solve those two problems,  we propose NaturalFinger, which generates natural query samples with generative adversarial networks (GANs)~\cite{GAN}. The intuition of our method is shown in Fig.~\ref{fig:intuition}. Our proposed NaturalFinger fingerprints the decision difference areas (purple areas) between positive models (stolen from the source model) and negative models (unrelated to the source model), rather than the decision boundary. Beneficial to this strategy, NaturalFinger is resistant to various model modifications. What's more, the utilization of GAN not only empowers us to generate more imperceptible samples, but also enables us to generate unrestricted samples to explore the decision boundary. Fig.~\ref{fig:demo} shows some interesting examples of our scheme. As shown in Fig.~\ref{fig:demo}, those samples contain some contentious features, which causes positive models and negative models to produce different labels. For example, Fig.~\ref{fig:demo}(a) is a bird with green color, positive models predict it as a bird while negative models predict it as a frog. 

However, simply applying GAN faces the following questions. 1) How to generate samples that fingerprint the decision difference areas? 2) How to generate natural query samples instead of low-quality images as we optimize the input noises? 3) How to reduce over-fit when the number of trained models is small? 

To address the first question, considering samples belonging to decision difference areas are predicted differently by positive models and negative models, we assign two different labels for them during optimization. For positive models, we assign the original labels predicted by most positive models. For negative models, to ease  the optimization, we assign them the closest labels to the original labels. In specific, we attack the negative models with a fast gradient sign method (FGSM)~\cite{FGSM} attack to obtain the adversarial labels. We then optimize the noises to ensure that all models can correctly predict the generated images. To handle the second question, we utilize the discriminator to constrain the generated samples by adding its loss to the total loss. To tackle the last question, inspired by data augmentation, we apply image transformation to the generated samples before feeding them into the model.

%
In summary, we propose a stealthy, robust, and effective model fingerprinting scheme to protect the IP of the model owner. The main contributions are summarized as follows:
\begin{itemize}[leftmargin=*]
	\item We propose NaturalFinger which generates natural and stealthy query samples with GAN. Besides, NaturalFinger fingerprints the decision difference areas instead of the decision boundary, which is more robust against model modification and is more effective against model extraction.
	
	\item To solve the challenges encountered by applying GAN, we propose three tricks including adversarial label, discriminator loss, and input transformation. Ablation experiments show that those tricks significantly improve the performance of NaturalFinger from 0.79 to 0.91 in ARUC.
	
	\item Experiments demonstrate that NaturalFinger is robust against four model modifications and is effective against two model extraction attacks. It attains a 0.91 ARUC value on the FingerBench dataset (154 models), exceeding the optimal baseline (MetaV~\cite{metav}) over 17\%. 
	
\end{itemize}


\section{Related Work}
In this section, we briefly overview the latest works on model fingerprinting. The goal of model fingerprinting is to construct a query set that can identify the ownership of the model. Based on the belief that a DNN classifier can be characterized by its decision boundary, many works generate data points that cross the decision boundary (adversarial examples) as the fingerprint. Cao \etal \shortcite{ipguard} propose IPGuard, which is the first method that utilizes adversarial examples to fingerprint the decision boundary. Then, Lukas \etal \shortcite{conferrableAdv} generate adversarial examples that only transfer from the source model to its stolen models as the fingerprint. Chang \etal \shortcite{DFfinger} use the DeepFool \cite{deepfool} algorithm to generate adversarial examples. Peng \etal \shortcite{UAPFinger} claim that local adversarial examples can only capture the local geometry, which may fail to be transferred to the suspect model due to decision boundary variation during the model extraction. Thus, they use universal adversarial perturbation (UAP) \cite{UAP} to fingerprint the model’s decision boundary. 

Some other methods generate noisy samples to fingerprint the model. Wang \etal \shortcite{wang2021intrinsic} use intrinsic examples as the fingerprint, which are generated by optimizing the noise to maximize the cross-entropy loss of a target label.  Yang \etal \shortcite{metafinger} fingerprint the noisy decision area of the source model by adding noises to natural samples. Pan \etal \shortcite{metav} leverage meta-training to optimize noisy inputs and use a meta-classifier to judge the ownership.

Whether the query samples are adversarial examples or noisy examples, they are unnatural images, which can be easily detected by the adversary. Unlike those unnatural samples, we generate imperceptible, natural samples with GAN.

\begin{figure*}[t] 
	\centering  

	\includegraphics[width=0.8\linewidth]{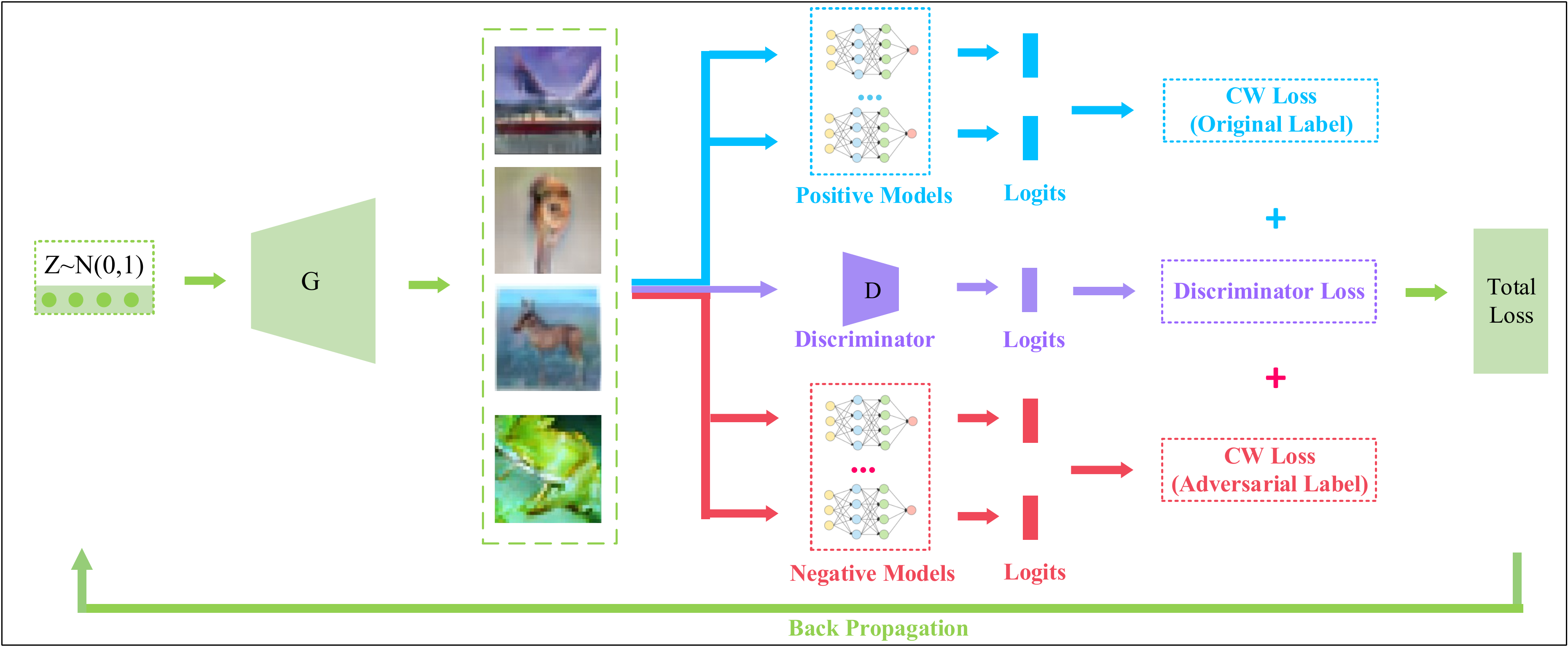}
	
	\caption{The illustration of our framework, which generates natural fingerprint with GAN. It assigns two different labels for positive models and negative models, then optimizes the generated images to ensure that all models are  correctly classified them.}
	\label{fig:framework}
\end{figure*}

\section{Method}

\subsection{Threat Model}
We consider a threat model which involves two parties: the model owner and the adversary. The model owner trains the source model, which is the model to be protected. The attacker has stolen the models from the source model by model extraction or illegal copy. 

The model owner aims to identify the stolen models as positive models and other unrelated models of honest third parties as negative models. The model owner has a white-box access to the source model and a black-box access to the suspect model. 

The attacker aims to escape the ownership verification while keeping a similar performance as the source model. We assume the adversary has some validation data to modify the model weights (model modification). We also assume the adversary can carry out two types of model extraction attacks: distillation attack requiring train data without labels and knockoff attack requiring other similar datasets. Finally, we suppose that the attacker can also adopt some techniques to detect or reject the abnormal query samples.

\begin{algorithm}[t]
	\caption{NaturalFinger}
	\textbf{Input:}
	Generator $\mathcal G$, Discriminator $\mathcal D$,
	Learning Rate $\eta$, Training Epoch $N_{e}$, Loss Control Coefficient $\lambda$, Positive Models $ \mathcal M_{p}$, \# Positive Models $ m$, Negative Models $\mathcal M_{n}$, \# Negative Models $ n$.\\
	\textbf{Output}: Query Set $Q$
	
	\begin{algorithmic}[1]
		
		\STATE $z = select\_initial\_samples(\mathcal G, \mathcal M_{p})$ 
		\STATE $Y_{original} = get\_original\_label(X,\mathcal M_{p})$ 
		\STATE $Y_{adv} = get\_adversarial\_label(X,\mathcal M_{n})$ 
		\FOR{$e=1$ to $N_{e}$}
		
		\STATE $X_{fake} = \mathcal G (z)$
		\STATE  $X = transform(X_{fake})$
		
		\STATE  $ L_{pos} =  \frac{1}{m}\sum\limits_{i = 1}^{m} {CW(\mathcal M_{p}^i(X), Y_{original})}$
		\STATE  $ L_{neg} =  \frac{1}{n}\sum\limits_{i = 1}^{n} {CW(\mathcal M_{n}^i(X), Y_{adv})}$
		
		\STATE  $L_{d} = HingeLoss(\mathcal D(X_{fake}), -1)$
		\STATE  $Loss = L_{pos} + L_{neg} + \lambda L_{d}$ 
		\STATE  $z = z - \eta \nabla Loss$
		\ENDFOR
		\STATE $Q = screen(X, Y_{original}, Y_{adv}, \mathcal M_{p}, \mathcal M_{n})$
		\STATE \textbf{return} $Q$
	\end{algorithmic}
	\label{alg:framework}
\end{algorithm}

\subsection{NaturalFinger Framework}
As shown in Fig.~\ref{fig:framework}, our method optimizes the input noises of GAN to ensure that the generated images are classified differently by positive models and negative models. As shown in the Algorithm \ref{alg:framework}, it mainly contains the following steps.

\textbf{Select Initial Samples (line 1).}
Given Gaussian noises to a GAN, it will generate images. However, we cannot guarantee the quality of those images. The low-quality images not only hinder the optimization but also are not stealthy.  Therefore, we first screen out the high-quality images as the initial samples. We believe that if an image can be correctly classified by all positive models, it is of high quality.

\textbf{Assigning Different Labels (lines 2 and 3).}
To fingerprint the decision difference areas, we need to generate the images that are predicted differently for positive models and negative models. Therefore, we assign two different labels for positive models and negative models during optimization. For positive models, we assign them the original labels predicted by them. For negative models, the simplest way is to randomly select the labels that are different from the original labels, but this may hinder  the optimization process. Different from this, we assign adversarial labels to negative models. In specific, we generate adversarial examples $X'$ by performing an untargeted FGSM attack on negative models for images $X$, then we take the predicted labels of adversarial examples $X'$ by most negative models as adversarial labels. 

\textbf{Optimize the Samples (line 4 $\sim$ 12).}
To reduce over-fit, we transform the generated samples before feeding them into the models. The input transformation also forces the generated images to include more robust features instead of noises. Then we compute the loss as:

\begin{equation}
\begin{aligned}
	Loss = L_{pos} + L_{neg} + \lambda L_{d} 
\end{aligned}
\label{eq:loss}
\end{equation}
This loss consists of three terms. The first term is the loss of all positive models as Eq.~\ref{eq:pos}, which encourages all positive models to predict the images $X$ as $Y_{pos}$.
\begin{equation}
\begin{aligned}
 L_{pos} =  \frac{1}{m}\sum\limits_{i = 1}^{m} {CW(p_i(X), Y_{pos})}
\end{aligned}
\label{eq:pos}
\end{equation}
where $CW(\cdot)$ denotes the Carlini-Wagner loss \cite{CW} as:
\begin{equation}
CW(Z, t) = \max (\max \{ Z_i:i \ne t\}  - Z_t, -k)
\end{equation}
where $Z$ denotes the inputs to the softmax function, also called logits. $t$ is the target label and the parameter $k$ encourages the GAN to generate samples that will be classified as class $t$ with high confidence.

The second term averages the loss of all negative models as Eq.~\ref{eq:neg}, which forces  all negative models to output the images $X$ as $Y_{neg}$.
\begin{equation}
\begin{aligned}
 L_{neg} =  \frac{1}{n}\sum\limits_{i = 1}^{n} {CW(n_i(X), Y_{neg})}
\end{aligned}
\label{eq:neg}
\end{equation}

The last term is the loss of discriminator $D$ that constrains the GAN to avoid generating unnatural images. In specific, we apply a discriminator to judge whether the generated images are fake or not. We apply the hinge loss as the discriminator loss:
\begin{equation}
\begin{aligned}
HingeLoss(\hat y, y) = \max (0,1 - y \cdot \hat y)
\end{aligned}
\label{eq:discriminator}
\end{equation}

Note that since the discriminator is already trained, it is not necessary to feed the real images. Besides, this loss only suits GANs that utilize hinge loss to train their discriminators such as SAGAN \cite{SAGAN} and ReACGAN \cite{ReACGAN}. For GANs that apply other loss functions such as wasserstein loss \cite{WGAN-GP}, we need to adjust this loss function according to their discriminator loss. 

Since this loss is fully derivative w.r.t. the latent $z$, we adopt off-the-shelf non-convex optimizers (e.g., SGD) for gradient-based optimization. When the number of trained models is small, it is possible to conduct back-propagation over the whole trained models. However, when the number of trained models is large, the optimization is resource-consuming. As an alternative, we split the trained models in batch, and we update the input noises for each batch just like mini-batch gradient descent.

\textbf{Screen the Samples (line 13).}
After optimization, not all samples are desired, so we screen out the successful samples from $X_{fake}$. Specifically, we transform the samples $X_{fake}$ with image transformation. If all positive models and all negative models correctly classified the transformed samples as $Y_{original}$ and $Y_{adv}$, respectively, we add the samples to the query set. 


\section{Experiments}

To demonstrate the effectiveness and robustness of our scheme, we run experiments against four model modifications and two model extraction attacks. To evaluate the stealthiness, we train a detector to detect the query samples. We compare our method with three advanced model fingerprinting methods IPGuard \cite{ipguard}, MetaV \cite{metav}, and MetaFinger \cite{metafinger} on CIFAR-10 \cite{CIFAR10} benchmark dataset. 

\subsection{Setup}
\textbf{Dataset Splitting.} We split the train data into two parts with the same number of samples for each classes.  One of the parts is used to train positive models and the other is used to train negative models. We also split the original test data into fine-tuning data and test data with the same number of samples for each classes. The fine-tuning data are used to fine-tune models and the test data are used to evaluate the model's  accuracy.

\textbf{Model Fingerprinting Benchmark Dataset.} Previous methods evaluate their methods with their own model dataset, which is not convenient for a fair comparison. To better evaluate all fingerprinting methods, we build a benchmark named FingerBench which includes 154 models on CIFAR-10 involving six attacks. To guarantee the variety of the benchmark, we use 15 state-of-the-art image classification model architectures to construct the benchmark. Tab.~\ref{tab:FingerBench} lists the construction of our model dataset. 

We use ResNet18 \cite{resnet} architecture as the source model and adopt six attacks to generate 63 positive models (including the source model). In specific, we attack the source model to generate 4, 9, 9, 10, 15, and 15 models with fine-tuning \cite{adi2018turning}, weight pruning \cite{pruning}, weight noising \cite{metafinger}, adversarial training \cite{wmrobustness}, distillation \cite{distillation}, and knockoff \cite{knockoff}, respectively. 

The construction of negative models is depended on whether the model architecture is the same as the source model. If the model architecture is the same as the source model (ResNet18), we run the same attacks as the source model which derives 63 mirror negative models. We term those negative models as mirror negative models because they undergo the same attacks as the positive models except the train data. For the other 14 model architectures, we train one model from scratch. We then steal a ResNet18 model with distillation for each model. The details of building the FingerBench model dataset can be found in supplementary material.

\begin{table}[t]
	\centering
	\scriptsize
	\caption{The construction of the FingerBench dataset.}
        \setlength\tabcolsep{2pt}
	\resizebox{\columnwidth}{!}{
	\begin{tabular}{lllp{4.5cm}}
		\toprule
		& Attack & \# & How to Generate \\ 
		\midrule
		\multirow{7}[11]{0.9cm}{Positive Model}  & Training & 1     & Train a ResNet18 model as the source model.\\
  	\cmidrule{2-4}     & Fine-Tuning & 4     & Fine-tune one model for each mode as in \cite{adi2018turning}.\\
		\cmidrule{2-4}          &Weight Pruning & 9     &  Prune the model weights with pruning rate $p$, then fine-tune it $p$ = 0.1, 0.2, ... , 0.9.  \\
		\cmidrule{2-4}          &Weight Noising & 9     & Perturb model weights with Gaussian noise, then fine-tune it.\\
		\cmidrule{2-4}          & Adversarial Training & 10    & Generate adversarial examples for $r$ percent of  the fine-tuning data, then fine-tune the model with the augmented dataset.  $r$ = 0.1, 0.2, ... , 1.\\
		\cmidrule{2-4}          & Distillation & 15    & Distill one model for each model architecture.      \\
		\cmidrule{2-4}          & Knockoff & 15    & Steal one model with Knockoff attack for each model architecture.  \\
		\midrule
		\multirow{3}[8]{0.9cm}{Negative Model} & Training & 14    & Train one model from scratch for each model architecture except the ResNet18 architecture.\\
		\cmidrule{2-4}          & Distillation & 14    & Distill one model for each model architecture except the ResNet18 architecture. \\
	
		\cmidrule{2-4}          & All Attacks & 63    & Perform the same attacks as the source model for ResNet18 architecture to generate mirror negative models.  \\
		\bottomrule
	\end{tabular}%
 }
	\label{tab:FingerBench}%
	
	%
	%
	
\end{table}%

\begin{figure*}[t]
	\centering  

	\setlength\tabcolsep{1pt}
	\begin{tabular}{cccc}
		\includegraphics[width=0.24\linewidth]{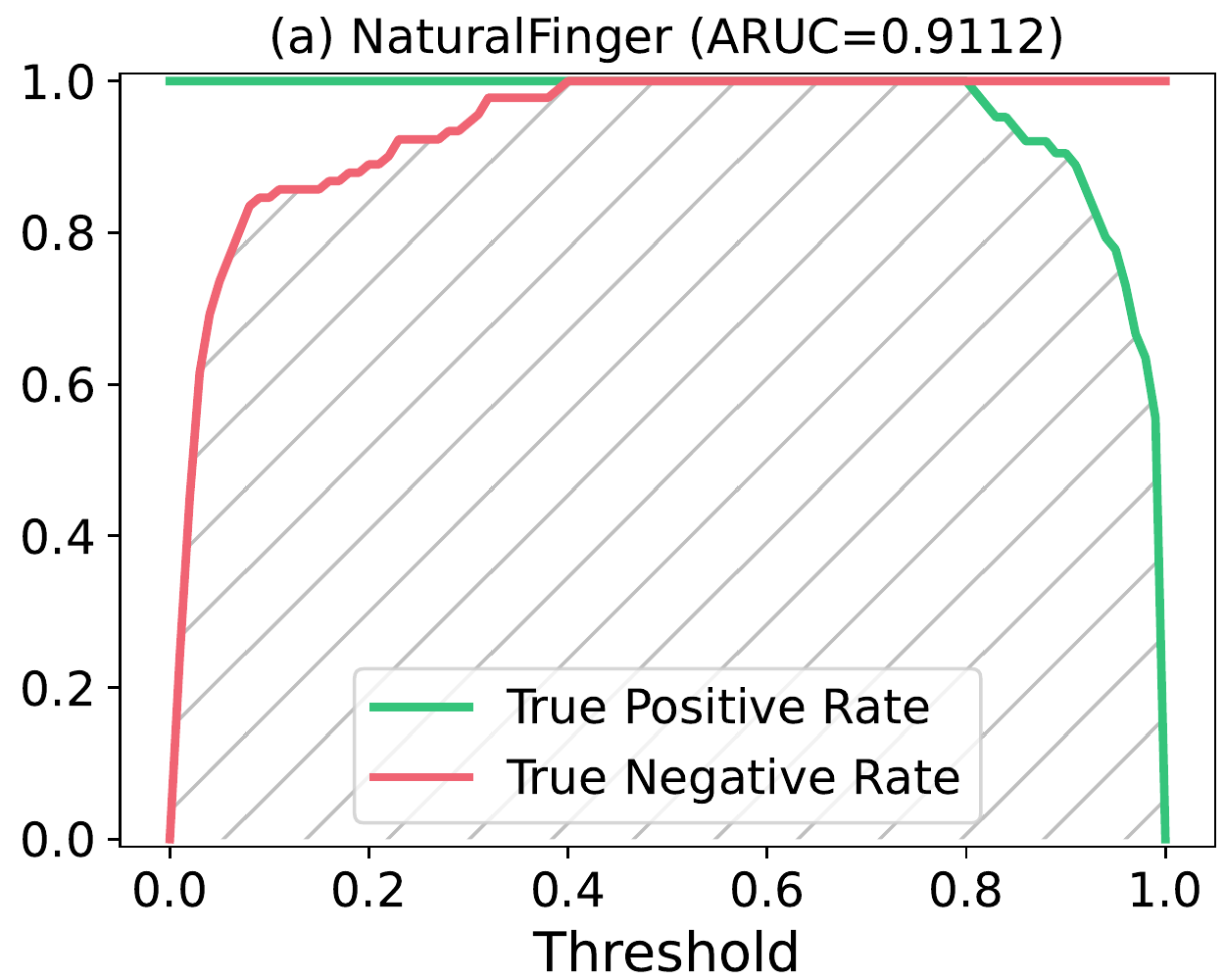} &
		\includegraphics[width=0.24\linewidth]{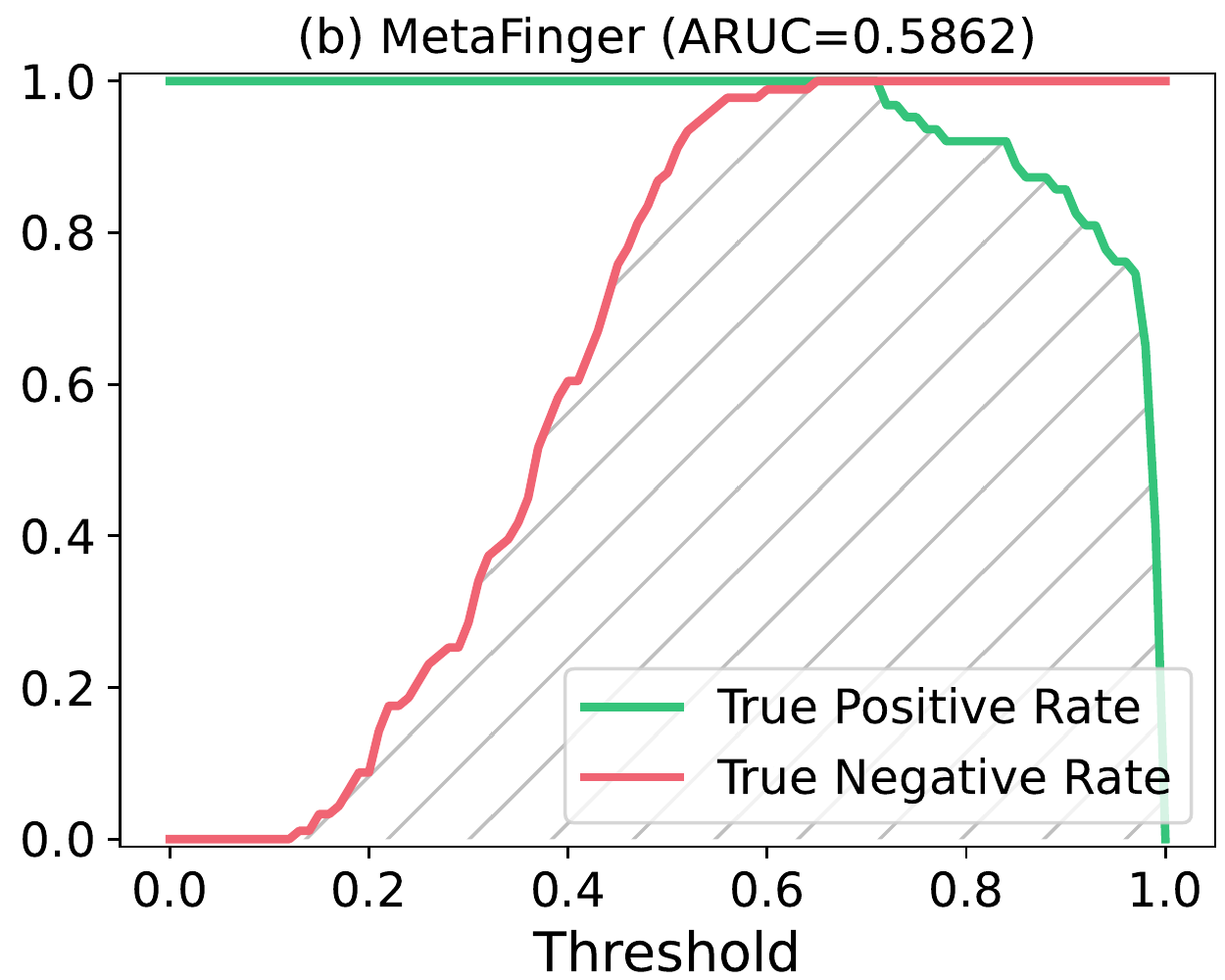} &
		\includegraphics[width=0.24\linewidth]{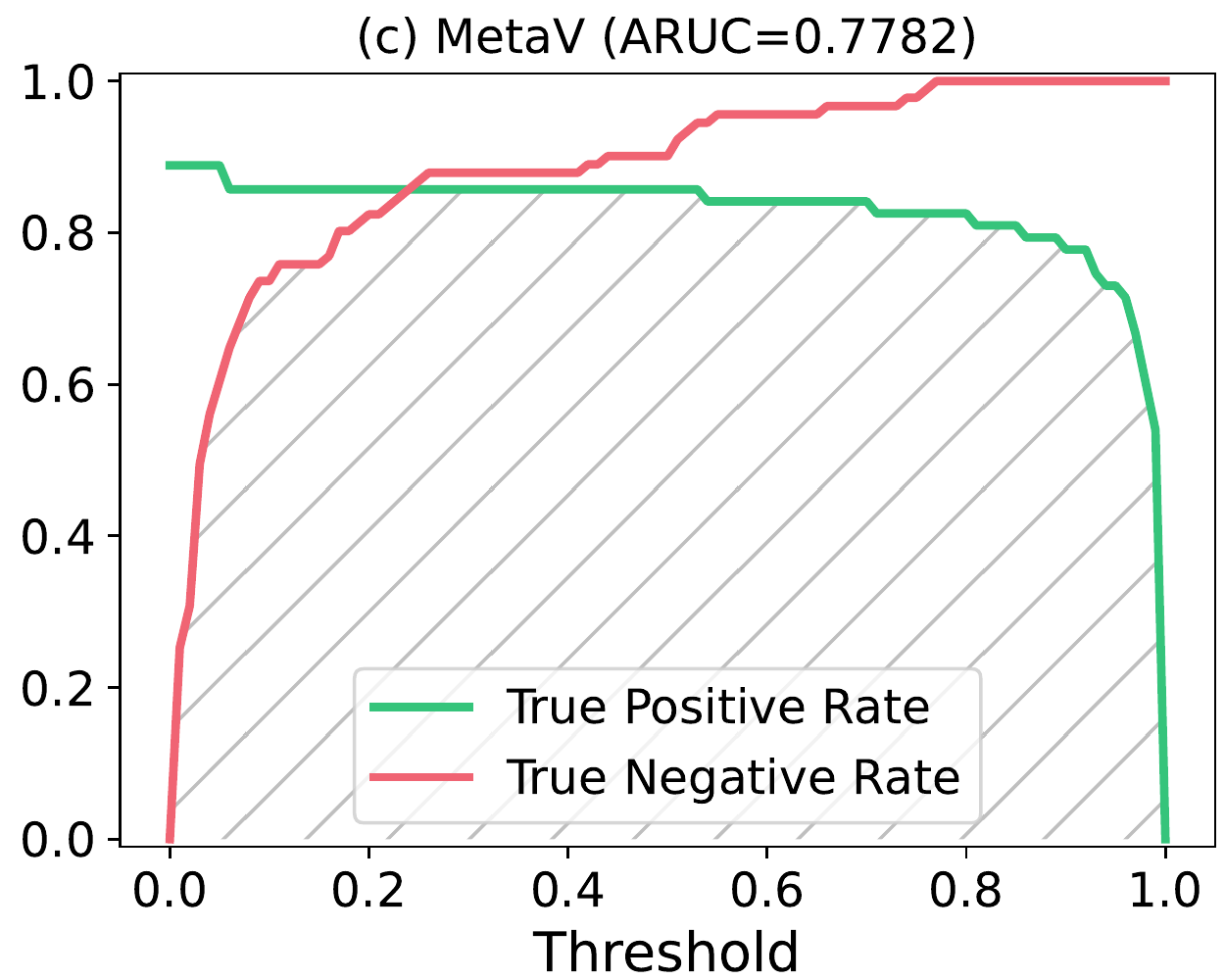}&
		\includegraphics[width=0.24\linewidth]{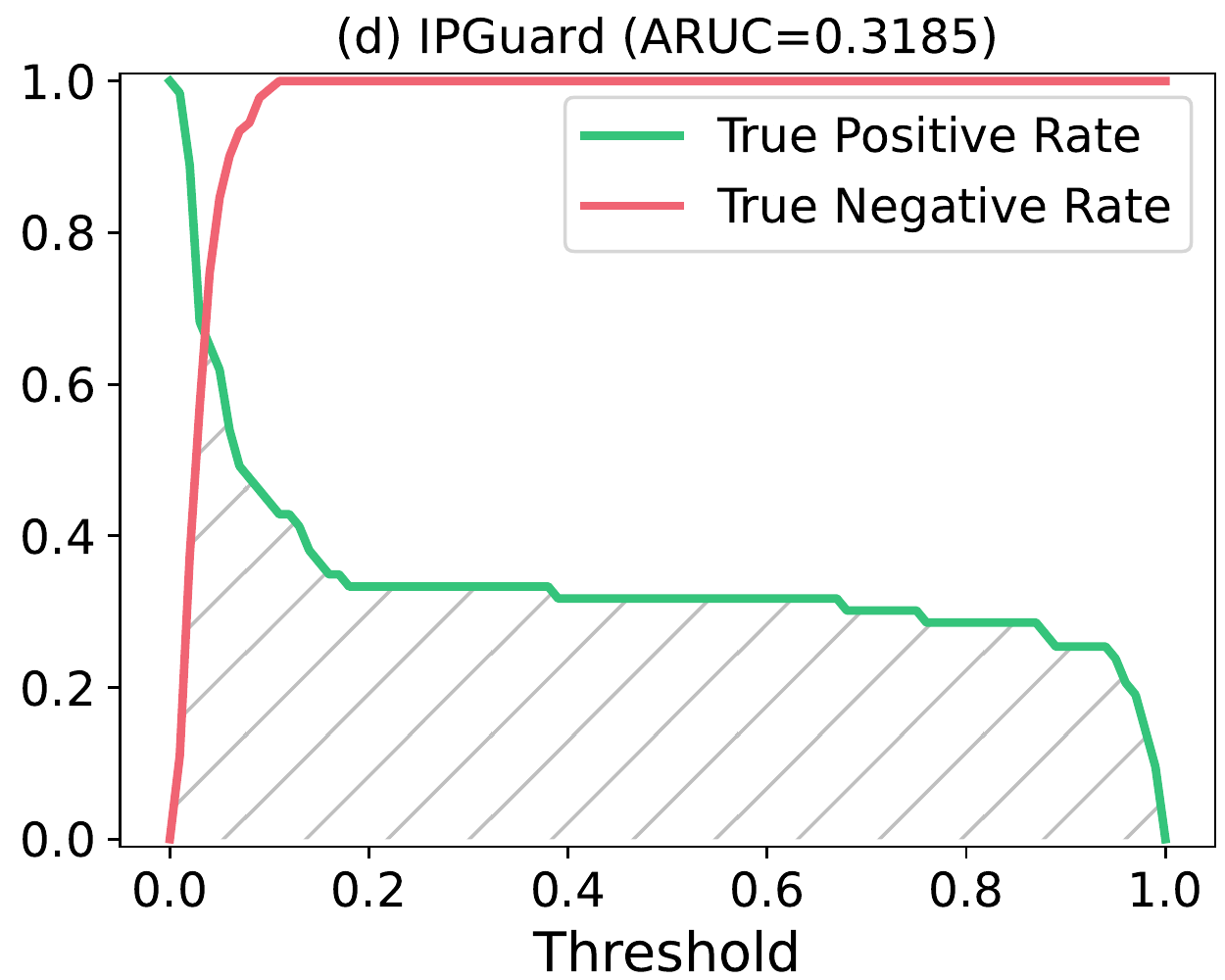}
	\end{tabular}
	\caption{The ARUC results of all methods on the FingerBench dataset.}
	
	\label{fig:cifar_aruc}
\end{figure*}

\begin{figure*}[t]
	\centering  

	\scriptsize
	\setlength\tabcolsep{1pt}
	\begin{tabular}{cccc}
		\includegraphics[width=0.245\linewidth]{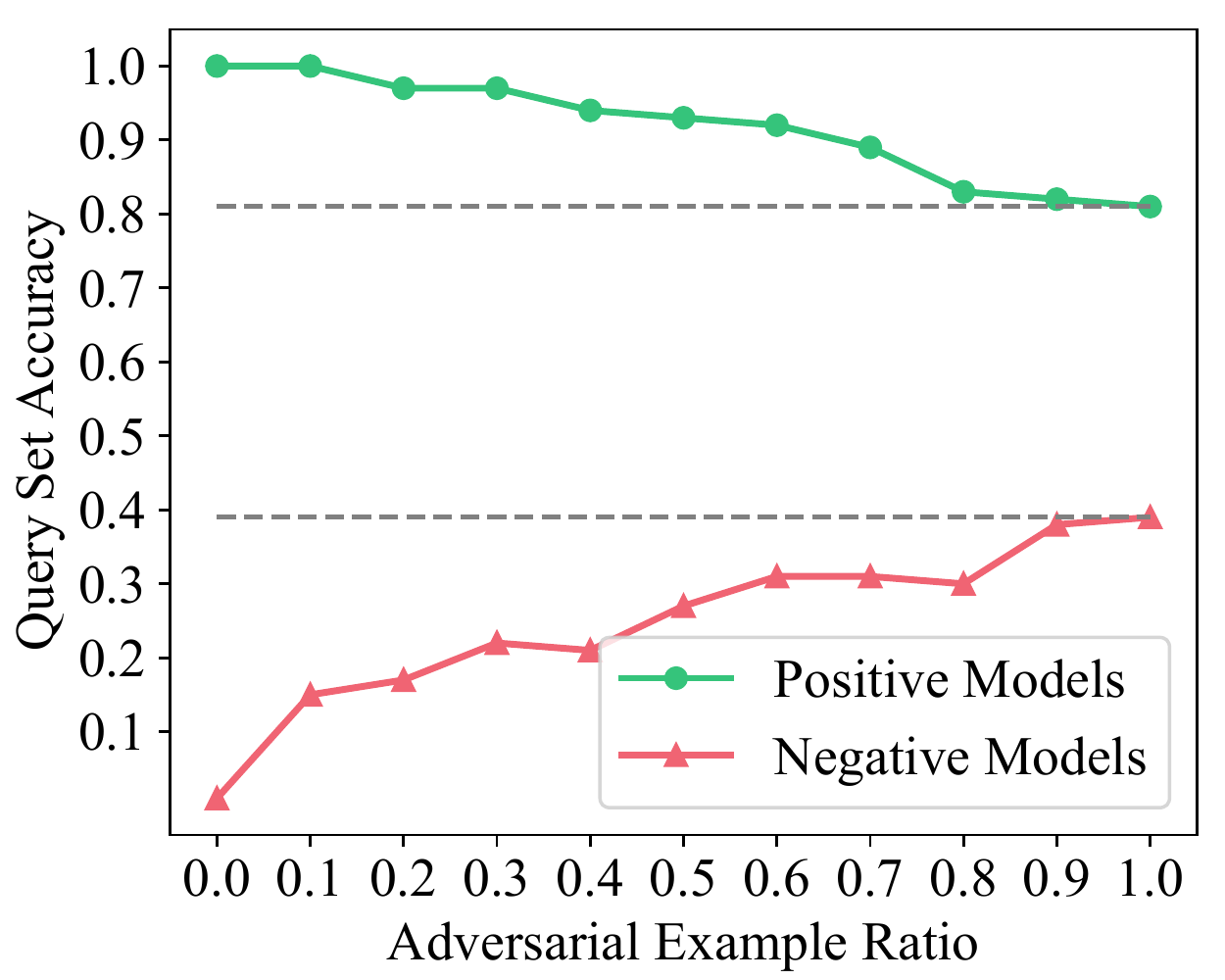} &
		\includegraphics[width=0.245\linewidth]{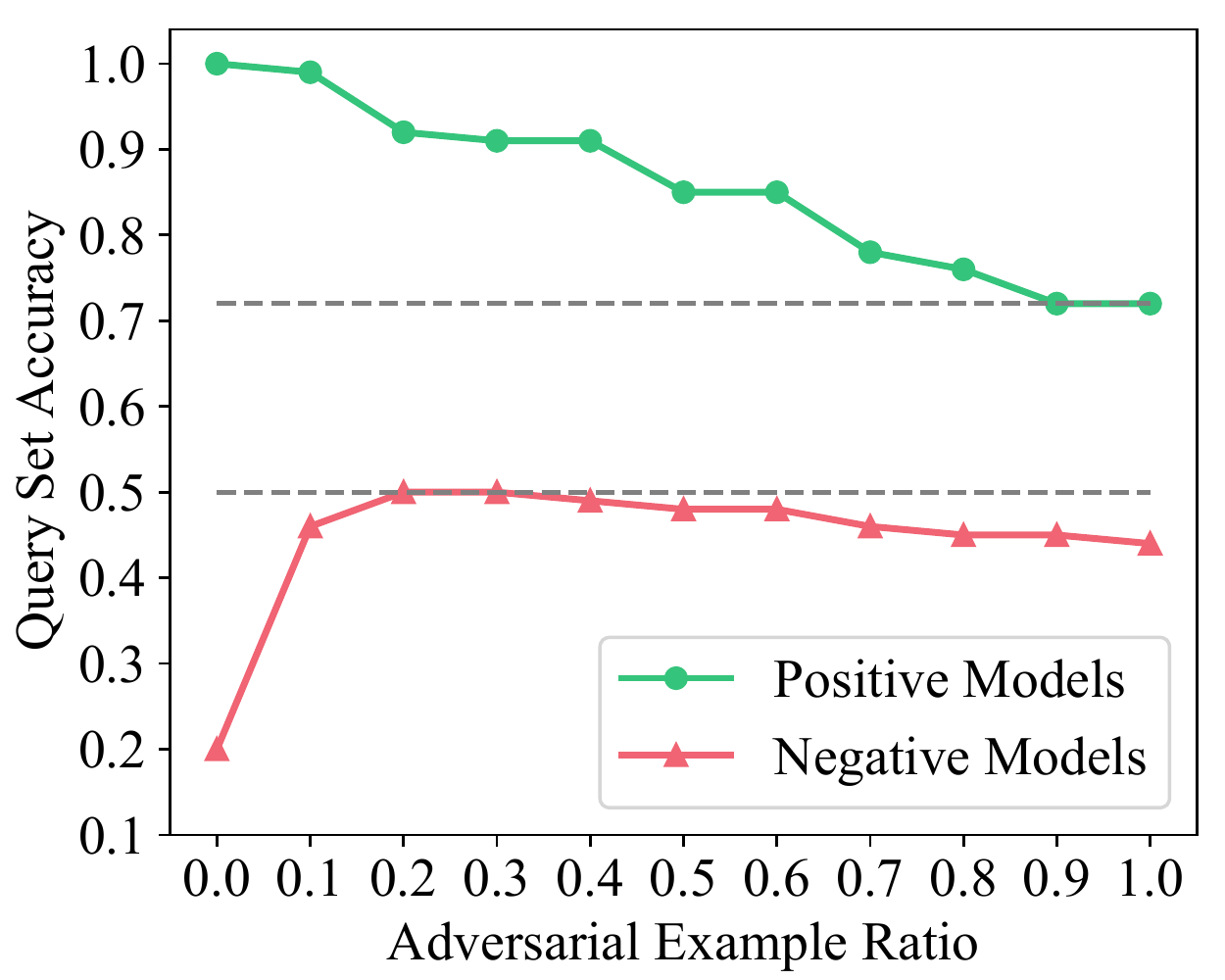} &
		\includegraphics[width=0.245\linewidth]{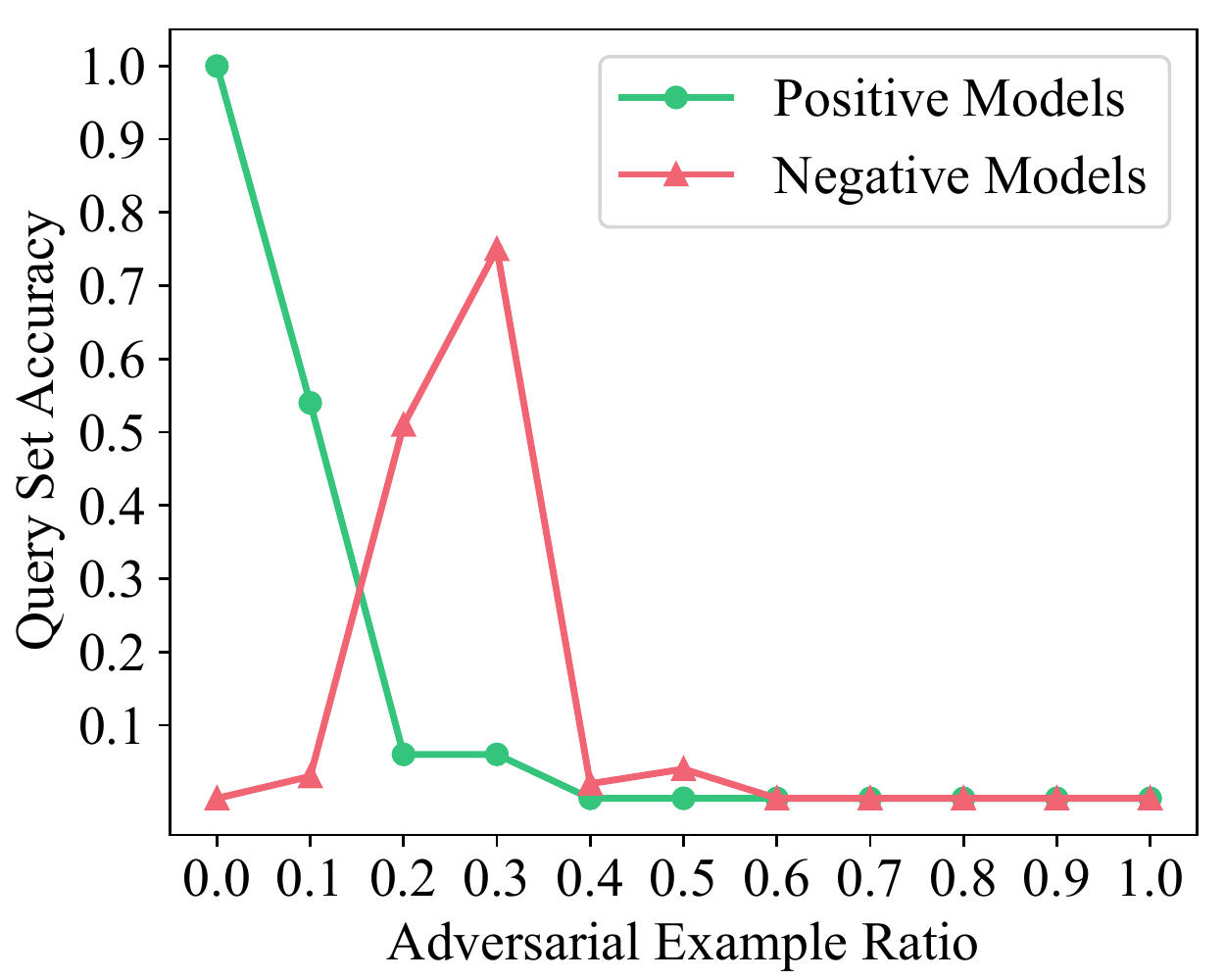}&
		\includegraphics[width=0.245\linewidth]{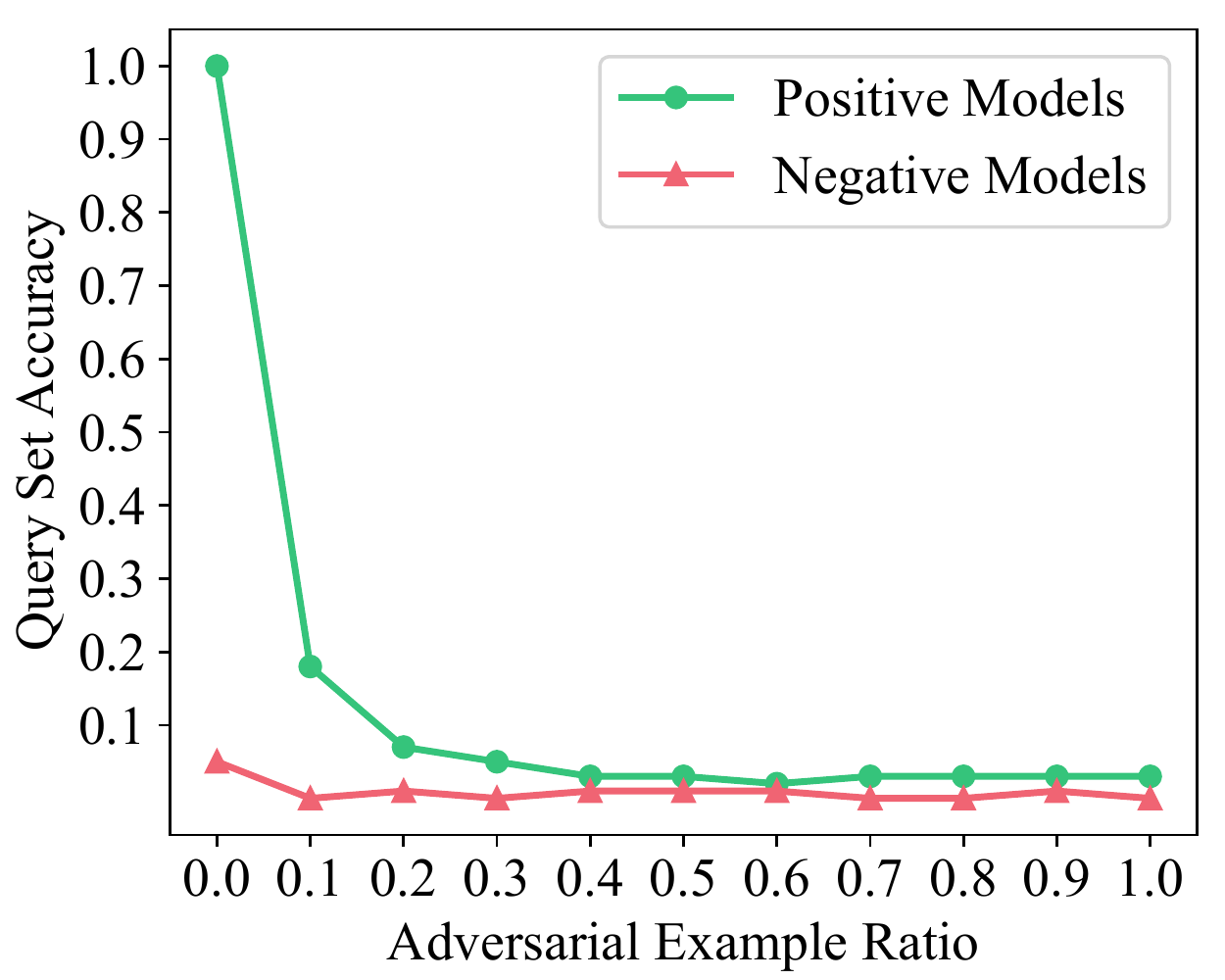} \cr
		(a) NaturalFinger &(b) MetaFinger & (c) MetaV & (d) IPGuard
		
	\end{tabular}
	\caption{The matching rate of positive models and corresponding mirror negative models under adversarial training for (a) NaturalFinger, (b) MetaFinger, (c) MetaV, and (d) IPGuard on CIFAR-10 dataset.}
	
	\label{fig:AT}
\end{figure*}
\begin{figure*}[t]
	\centering  

	\scriptsize
	\begin{tabular}{ccc}
		\includegraphics[width=0.31\linewidth]{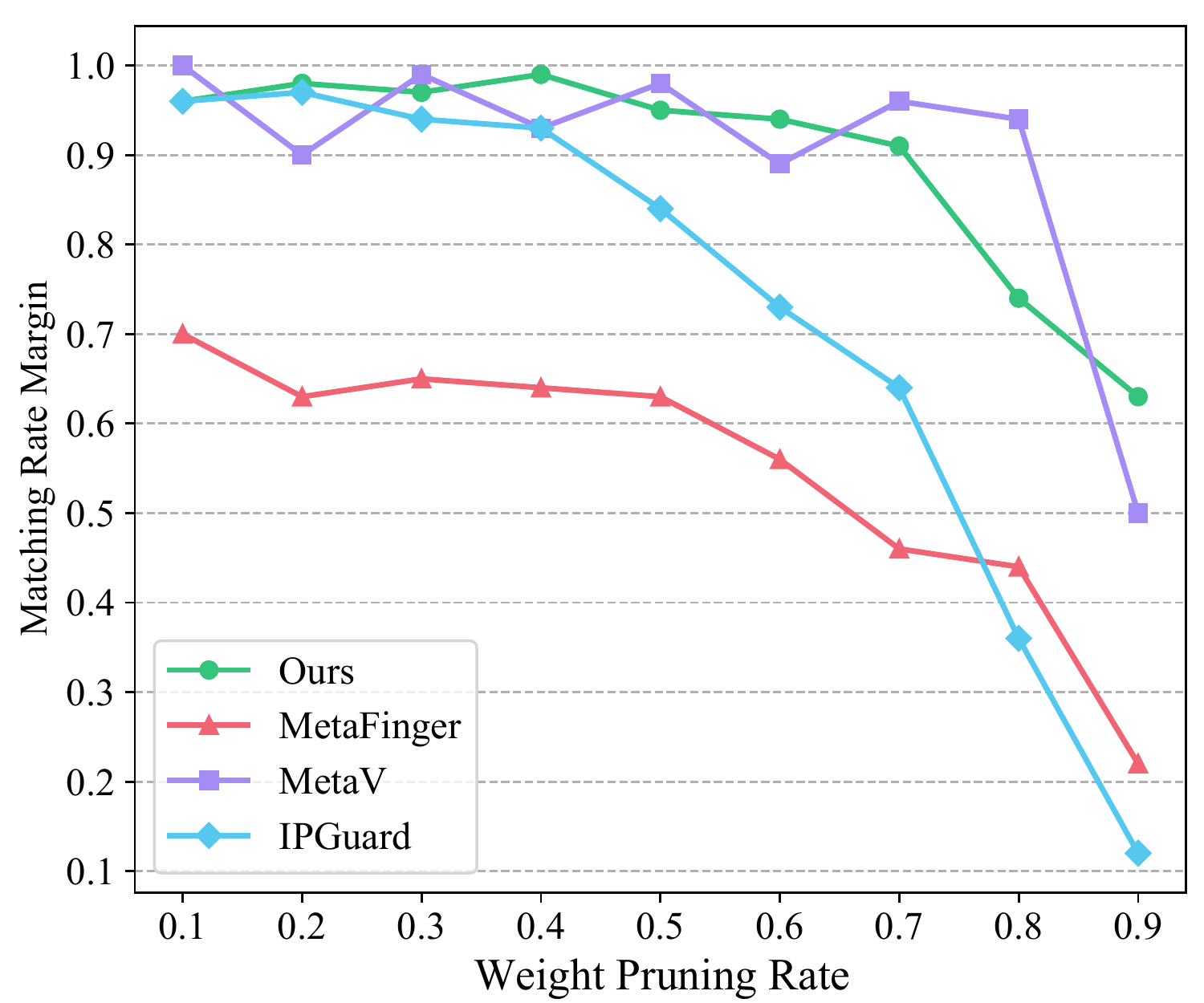} &
		\includegraphics[width=0.31\linewidth]{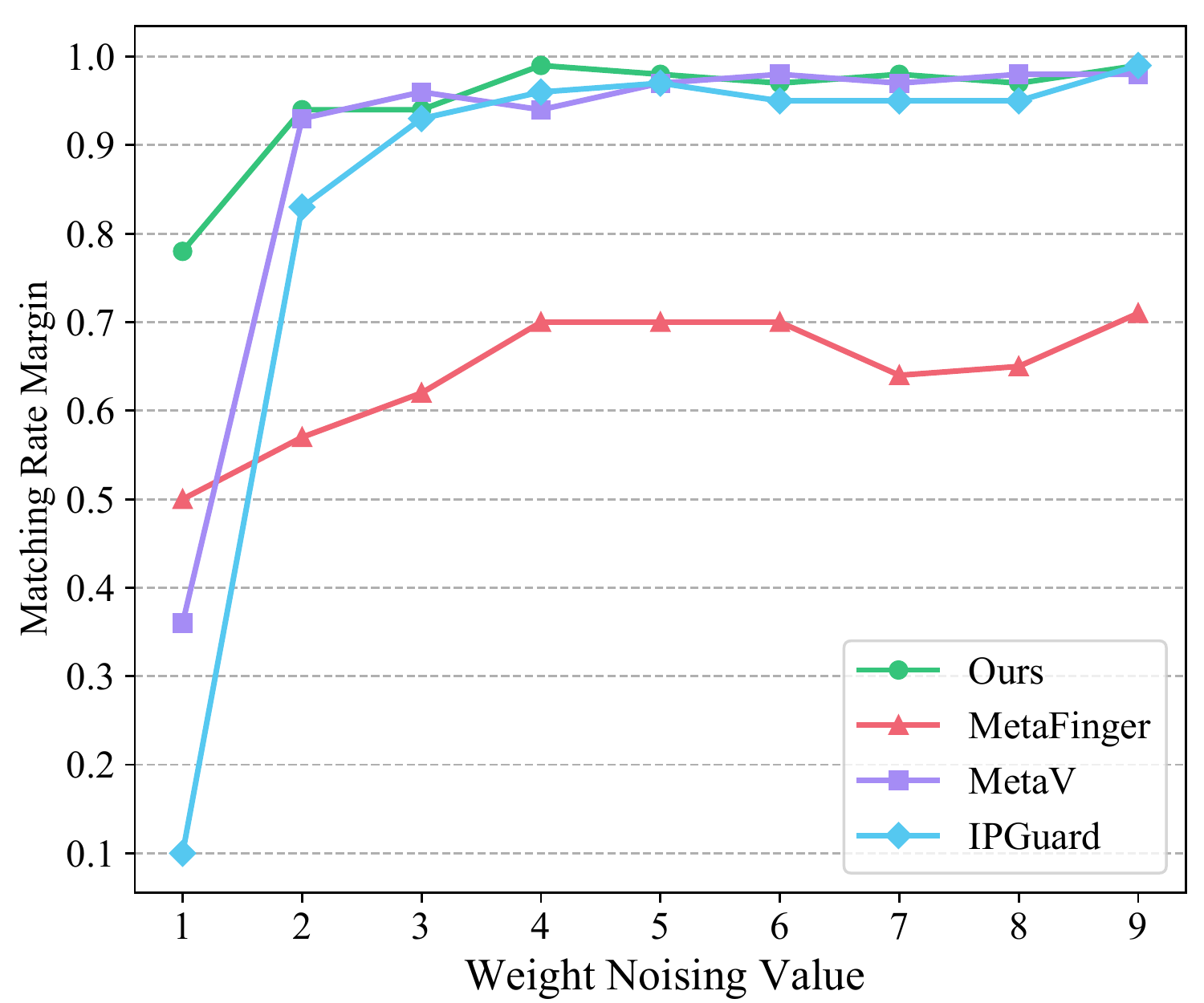} &
		\includegraphics[width=0.31\linewidth]{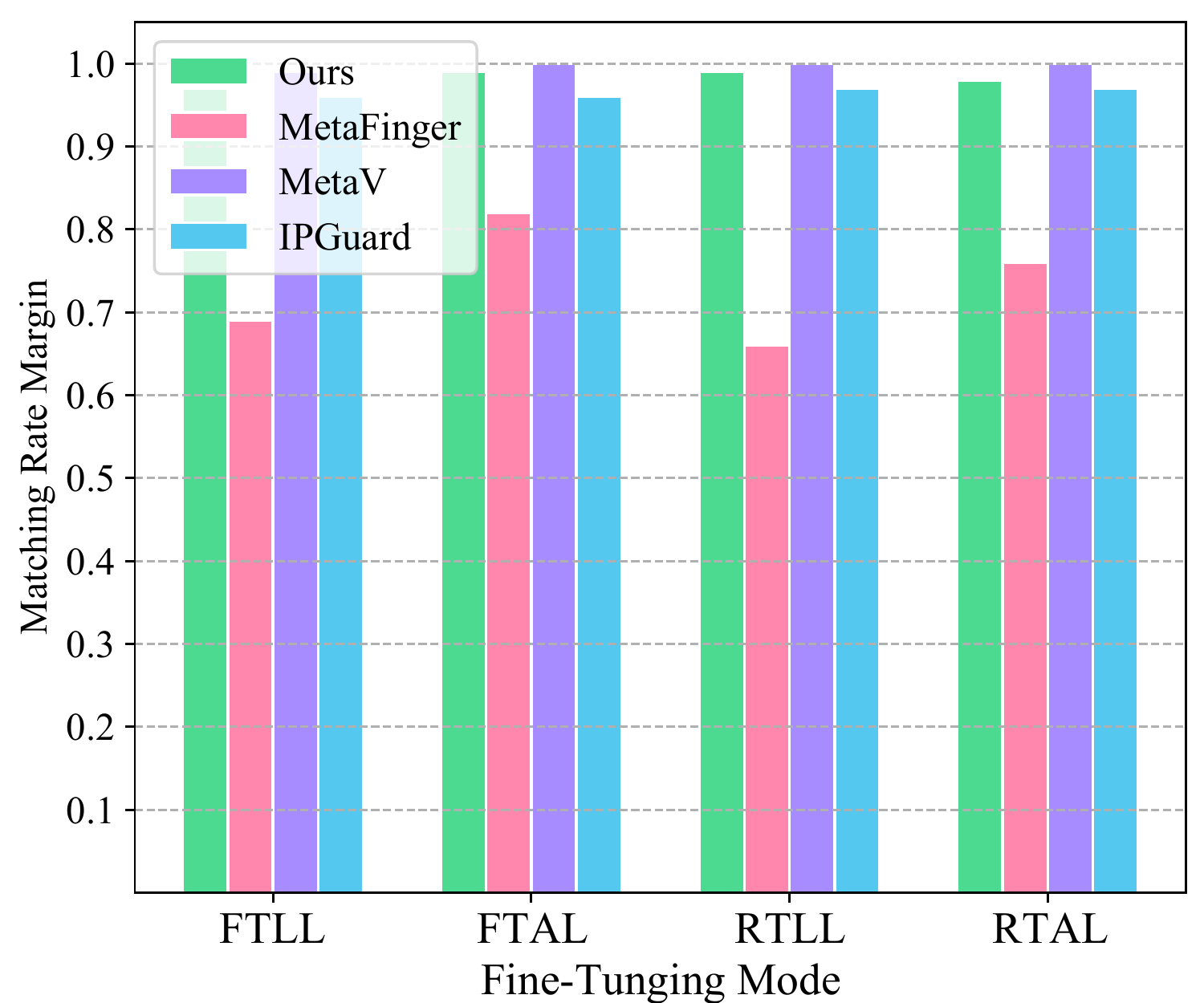} \cr
		(a) Weight Pruning & (b) Weight Noising & (c) Fine-Tuning
	\end{tabular}
	\caption{The matching rate margin under (a) Weight Pruning, (b) Weight Noising, and  (c) Fine-Tuning on CIFAR-10 dataset. The matching rate margin is computed by subtracting the matching rate of a mirror negative model from the matching rate of a corresponding positive model. This negative model is exactly the same as the positive model except for the train data.}
	
	\label{fig:cifar_model_modification}
\end{figure*}

\textbf{Evaluation Metrics.}
In the evaluation, we adopt the matching rate and area under the robustness-uniqueness curves (ARUC) metrics to evaluate all methods as done in \cite{ipguard}. Besides, we also introduce the matching rate margin to better compare the fingerprinting schemes.




(3) ARUC. ARUC measures the area of the intersection region under the robustness and uniqueness curves as the matching rate threshold increases from 0 to 1. Robustness, also called true positives rate (TPR), measures the proportion of positive suspect models also classified as positive. Uniqueness or true negative rate (TNR) measures the proportion of negative suspect models also classified as negative. This metric is used to evaluate the performance on the model dataset.

\textbf{Hyper-Parameters.} We select eight negative models and eight positives models from InceptionV3 \cite{InceptionV3}, ResNet18, ResNet34 \cite{resnet}, VGG13\_BN \cite{VGG}, and VGG16\_BN \cite{VGG} model architectures as the trained models. Among the trained models, some of the models are distilled and others are trained from scratch. We adopt SGD with a learning rate of 0.5 as the optimizer and the number of iterations as 1000.  We adopt SAGAN \cite{SAGAN} to generate samples and its discriminator to calculate the discriminator loss. Besides, we set the loss control coefficient $\lambda$ to 0.5. For a fair comparison, all fingerprinting schemes use the same trained models and  the number of fingerprint examples is set to 100.

\begin{figure*}[t]
	\centering  

	\includegraphics[width=\linewidth]{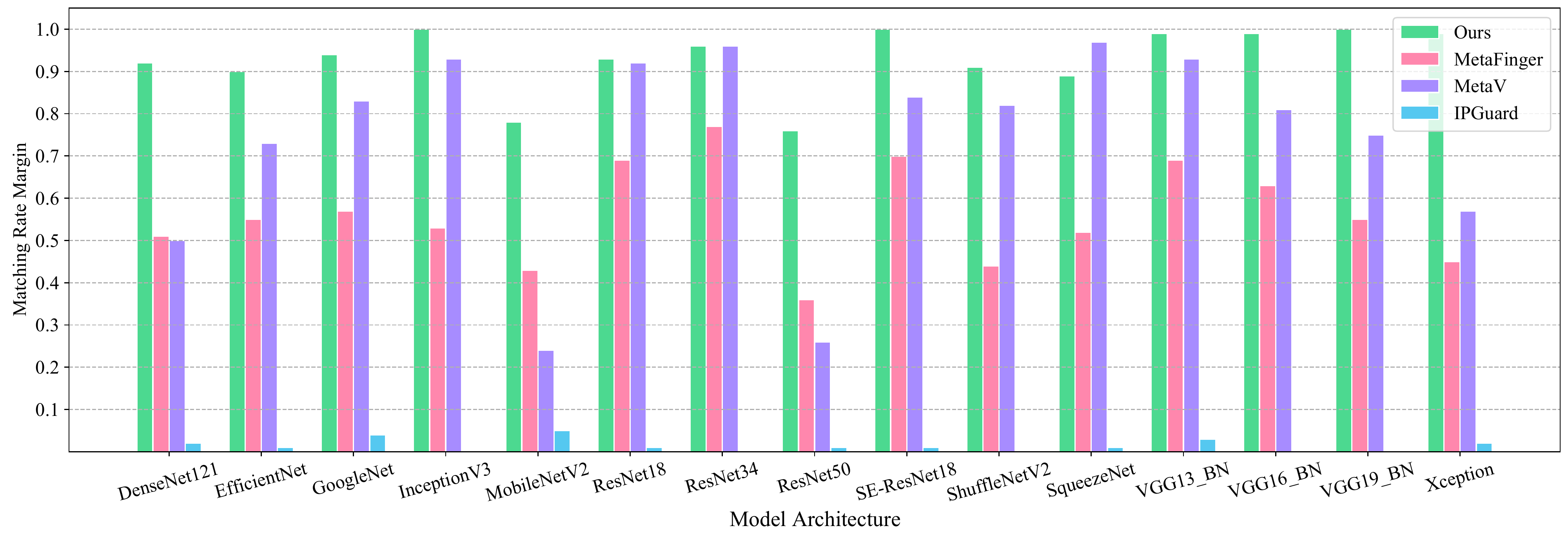}
	\caption{The matching rate margin of all methods on various model architectures against distillation attack.}
	
	\label{fig:distillation}
\end{figure*}

	
	

\subsection{Overall Results on the FingerBench Benchmark}
In this part, we compare the performance of NaturalFinger with three state-of-the-art model fingerprinting methods on the FingerBench dataset. Fig.~\ref{fig:cifar_aruc} shows the ARUC results of all methods. As Fig.~\ref{fig:cifar_aruc} shows, our proposed NaturalFinger achieves the highest ARUC value (0.9112), which exceeds the second highest method MetaV over 17\% (0.1330). We also notice that NaturalFinger has the largest threshold range that ensures the true positive rate and true negative rate simultaneously achieve 100\%.

\subsection{Robustness against Model Modification}
After stealing the model, the adversary may apply some model modifications to alter the model weights such as fine-tuning. Thus, we evaluate the robustness of all methods against model modifications including adversarial training, weight pruning, weight noising, and fine-tuning.

\textbf{Adversarial Training.} Fig.~\ref{fig:AT} shows the matching rate of positive models and corresponding mirror negative models when augmenting different proportions of adversarial examples for adversarial training. From Fig.~\ref{fig:AT} (c) and (d), we find that, for MetaV and IPGuard, the matching rate of positive models dramatically down to zero. The reason is that their samples are constructed from adversarial examples or noisy examples, positive models will correctly classify those samples when positive models become more robust. From Fig.~\ref{fig:AT} (a) and (b), we observe that as the increasing of adversarial examples, the matching rate of positive models gradually decreases and the matching rate of negative models gradually increases for NaturalFinger and MetaFinger. But since the samples of our approach and MetaFinger are not adversarial examples, the matching rate of positive models only slightly decrease. We assume this is because adversarial training changes the decision boundary and degrades the model performance.

\textbf{Weight Pruning.} As shown in Fig.~\ref{fig:cifar_model_modification} (b), the matching rate margin of all methods gradually declines as the weight pruning rate increases, since the large weight pruning rate downgrades the model performance. Even so, NaturalFinger and MetaV still achieve about 90\% matching rate margin when the weight pruning rate is less than or equal to 0.7.

\textbf{Weight Noising.} Fig.~\ref{fig:cifar_model_modification} (c) shows the results under different weight noising values. The smaller the value, the stronger the perturbation strength. All methods except MetaFinger obtain similar results when the weight noising value is great than 2. Only our approach still achieves over 75\% result when the weight noising value is set to 1.

\textbf{Fine-tuning.} Fig.~\ref{fig:cifar_model_modification} (a) presents the results of fine-tuning on CIFAR-10. We observe that all methods except MetaFinger can achieve over 95\% matching rate margin regardless of the fine-tuning mode. 



\subsection{Effectiveness against Model Extraction}
In addition to obtaining the model from an illegal copy, the adversary can also steal models by querying the source model. So we evaluate the performance of all methods against two model extraction attacks including distillation and knockoff attacks. As shown in Fig.~\ref{fig:distillation}, even though NaturalFinger only selects InceptionV3, ResNet18, ResNet34, VGG13\_BN, and VGG16\_BN as the trained models, it can still achieve excellent results on other model architectures, which shows great potential for unknown model architectures. For example, NaturalFinger achieves nearly 100\% matching rate margin on SE-ResNet18~\cite{seresnet}, VGG19\_BN \cite{VGG}, and Xception~\cite{xception}, which even attain better results than the trained models.  Although MetaV shows somewhat transferability, it cannot achieve similar results on unseen model architectures. The results of all methods against knockoff attack is similar to distillation attack, which can be found in supplementary material. 


\subsection{Stealthiness}
\begin{figure}[t]
	\centering  

	\setlength\tabcolsep{1pt}
	\scriptsize  
	\begin{tabular}{cccc}
		\includegraphics[width=0.24\linewidth]{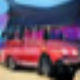} &
		\includegraphics[width=0.24\linewidth]{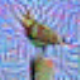} &
		\includegraphics[width=0.24\linewidth]{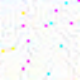}&
		\includegraphics[width=0.24\linewidth]{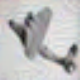} \cr
		
		(a) NaturalFinger & (b) MetaFinger & (c) MetaV & (d) IPGuard
	\end{tabular}
	\caption{The query samples of all evaluated methods.}
	
	\label{fig:all_demo}
\end{figure}

To find out the ownership of the suspicious model, fingerprinting schemes usually query the model in black-box setting. Therefore, the query samples should be imperceptible and undetectable. Otherwise, those samples may be detected and rejected by the adversary, hindering the verification process. Fig.~\ref{fig:all_demo} shows the query samples of all methods. Among all methods, NaturalFinger generates the most natural images in human eyes.

Far beyond human observation, we also train a nine layers convolutional neural network (CNN) to evaluate the stealthiness of the query samples. In specific, we assume the adversary does not have the knowledge of query samples. But we assume the adversary has 5k fine-tuning data to train the detector. We then generate 5k noise samples with Gaussian noise and 5k adversarial examples with FGSM attack. We apply Adam \cite{adam} with a learning rate of 0.001 for 20 epochs to train the CNN. Tab.~\ref{tab:stealthiness}  lists the detection results (average over three times). We can observe that NaturalFinger generates the most undetectable query samples since it generates natural images with GAN. Moreover, the detection rate of MetaV is also low, because its samples are empty images with slight noises, which does not obey the data distribution of train data. What's more, the detector can detect the samples of MetaFinger and IPGuard with high accuracy because their samples are constructed by adding noises to normal images, which is similar to the train data.

\begin{table}[t]
	\centering
	\caption{The detection rate of query samples for all methods.}
	\renewcommand{\arraystretch}{0.9}
	\resizebox{\columnwidth}{!}{
	\begin{tabular}{ccccc}
		\toprule
		Method& NaturalFinger & MetaFinger & MetaV & IPGuard \\
		\midrule
		Detection Rate  & 0.003 &   1.00    &  0.48     &  0.85    \\
		\bottomrule
	\end{tabular}%
	}
	\label{tab:stealthiness}%
\end{table}%

\begin{table}[t]
	\centering
	\caption{The ARUC results of NaturalFinger on the FingerBench when disabling some tricks.}
	\setlength{\aboverulesep}{0pt}
	\setlength{\belowrulesep}{0pt}
	
	\renewcommand{\arraystretch}{1.2}
	\resizebox{\columnwidth}{!}{

	\begin{tabular}{ccc|c}
		\toprule
		Adversarial Label & Input Transformation & Discriminator Loss & ARUC\\
		\midrule
		-&   -    &   -    & 0.79 \\ \midrule
		\checkmark&   -    &   -    &  0.84 \\
		-&       \checkmark&    -   & 0.86 \\
		-&   -    &      \checkmark&  0.80 \\
		\checkmark&       \checkmark&  -     & 0.87 \\
		\checkmark&    -   &       \checkmark& 0.86 \\
		-&       \checkmark&      \checkmark& 0.86 \\  \midrule
		\checkmark&       \checkmark&       \checkmark& \textbf{ 0.91} \\
		\bottomrule
	\end{tabular}%

	}
	\label{tab:ablation_tricks}%

\end{table}%
\begin{figure}[t]
	\centering  

	\includegraphics[width=0.7\linewidth]{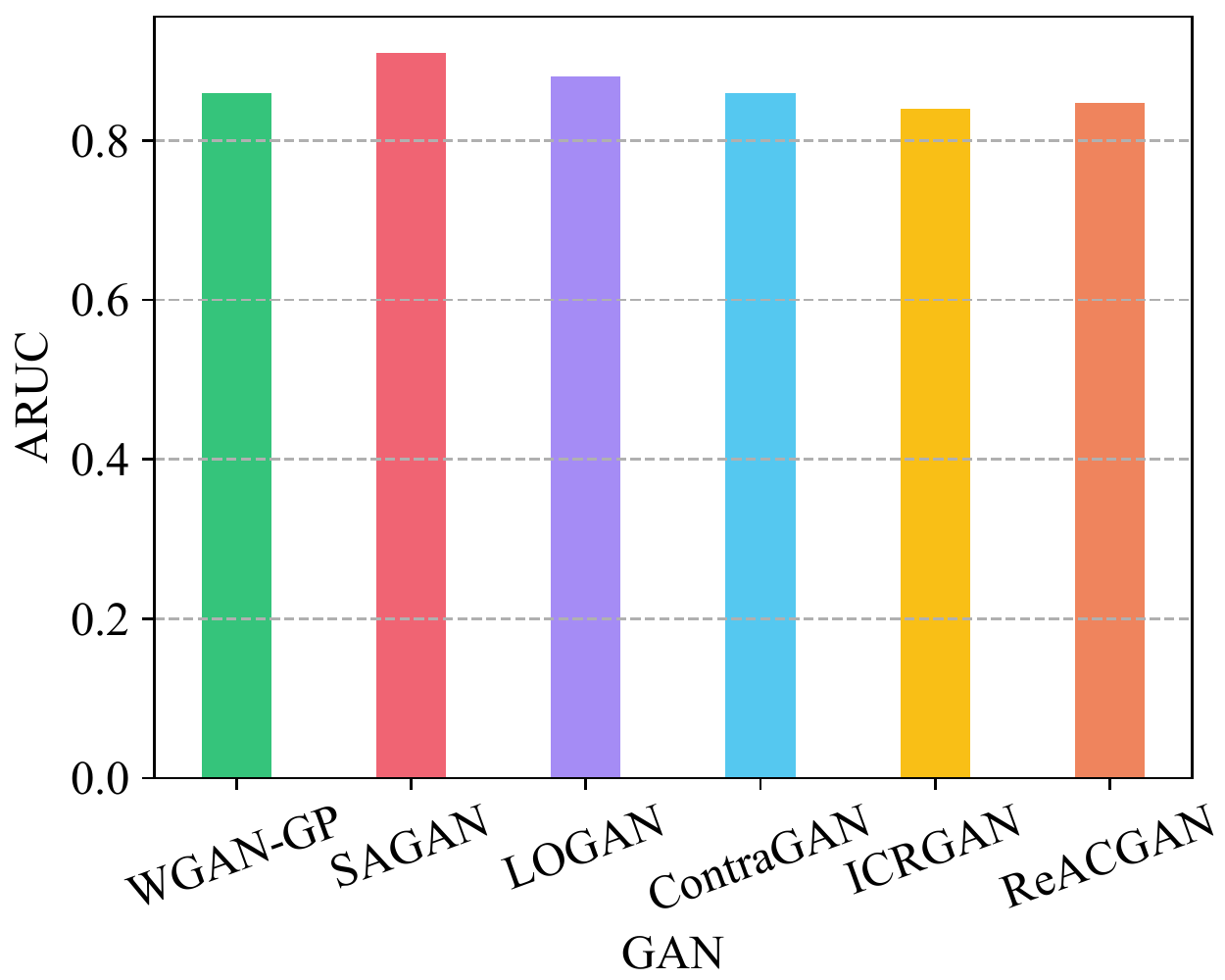}
	\caption{The ARUC of NaturalFinger on  the FingerBench when replacing SAGAN with other GANs.}
	
	\label{fig:GAN}
\end{figure}

\subsection{Ablation Study}

\textbf{The Effectiveness of Proposed Tricks.}
To investigate whether our proposed tricks improve the performance of NaturalFinger, we ablate those tricks to compare the results. Tab.~\ref{tab:ablation_tricks} lists the results when ablating different tricks. From this table, we can observe that our baseline can only achieve a 0.79 ARUC value and any one of the three tricks can improve the performance. Among those three tricks, input transformation and adversarial label significantly improve the results over baseline. Even though the discriminator loss barely improves the performance, it encourages the GAN to generate more natural images. By combining all the tricks, NaturalFinger achieves the highest result, improving the performance of the baseline over 15\%. 

\textbf{The Impact of GAN.}
To figure out whether the type of GAN significantly impacts the performance of NaturalFinger, we replace SAGAN with other five different GANs and keep all other hyper-parameters unchanging. In specific, to ensure the diversity of GAN, we select WGAN-GP \cite{WGAN-GP}, LOGAN \cite{LOGAN}, ContraGAN \cite{ContraGAN}, ICRGAN \cite{ICRGAN}, and ReACGAN \cite{ReACGAN}, which covers GAN from 2017 to 2021. 
From Fig.~\ref{fig:GAN}, we can see that although SAGAN (Ours) achieves the highest ARUC value, other GANs also obtain similar results. Considering that the hyper-parameters are adjusted for SAGAN, other GANs may attain better performance by fine-tuning the hyper-parameters. Therefore, we can believe that the type of GAN does not impact a huge influence on NaturalFinger. 

\section{Conclusion}
In this paper, we propose a stealthy, robust, and effective DNN fingerprinting scheme called NaturalFinger, which fingerprints the decision difference areas with natural images generated by GAN. To fingerprint these areas, we assign two different labels for negative models and positive models, then optimize the generated images to ensure all models correctly recognize the images. Besides, we also propose three ticks including adversarial label, discriminator loss, and input transformation to address the challenges faced by simply applying GAN. Experimental results demonstrate the advantages of NaturalFinger in the following aspects: 1) Robustness. NaturalFinger achieves similar or better results against four model modifications including adversarial training; 2) Effectiveness. NaturalFinger obtains over 70\% matching rate margin for all 15 different model architectures against two model extraction attacks; 3) Stealthiness. The query samples of NaturalFinger not only is imperceptible in human eyes, but also is hard to detect for a trained detector (0.3\% detection rate).


\bibliographystyle{named}
\bibliography{ijcai23}

\end{document}


\maketitle

\begin{figure*}[!htbp]
	\centering  

	\includegraphics[width=\linewidth]{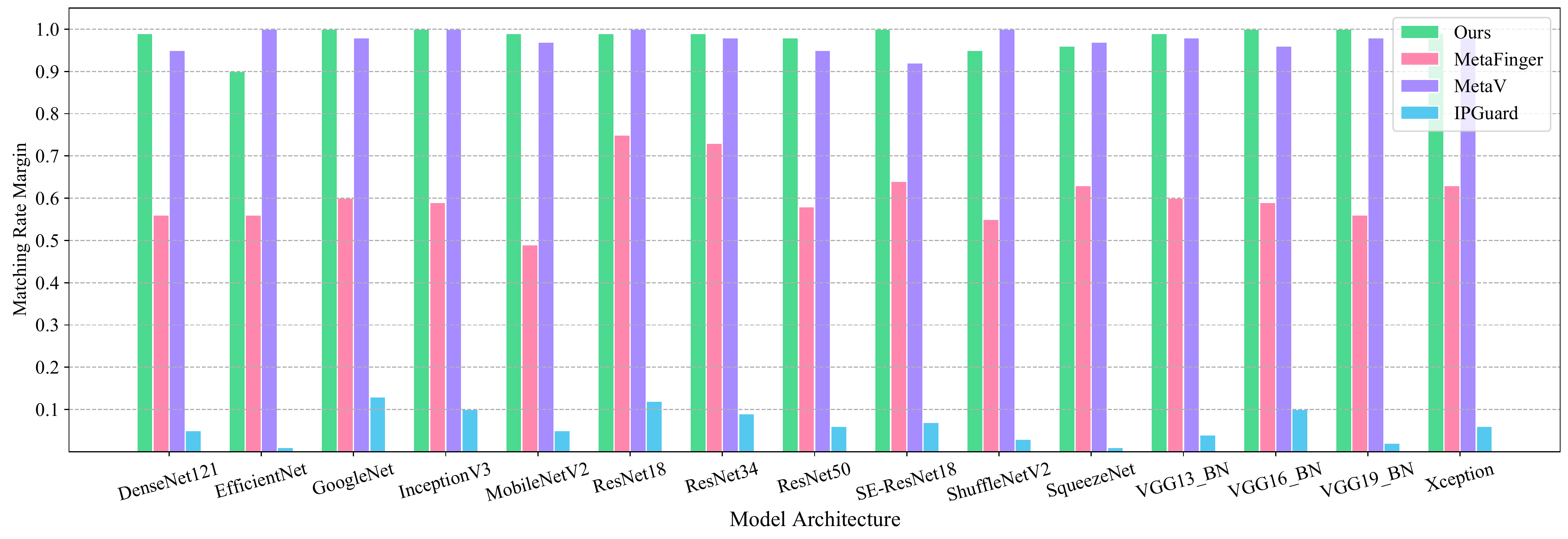}
	\caption{The matching rate margin of all methods on various model architectures against knockoff attack.}
	
	\label{fig:knockoff}
\end{figure*}
\section{FingerBench Dataset}
To better evaluate all fingerprinting methods, we build a benchmark named FingerBench which includes 154 models on CIFAR-10 involving six attacks. The following is the details of the attacks adopted in this paper to build the FingerBench model dataset.




We fine-tune the model 10 epochs with SGD optimizer with a learning rate of 0.005.

\textbf{Fine-tuning.} we fine-tune the model with fine-tune data in four modes as done in~\cite{adi2018turning}. 
\begin{itemize}
	\item Fine-Tune Last Layer (FTLL): Fine-Tune the last layer in the model and fix all other layers.
	\item Fine-Tune All Layers (FTAL): Fine-Tune all layers in the model.
	\item Re-Train Last Layer (RTLL): Re-initialize the weights of the last layer and only fine-tune the last layer.
	\item Re-Train All Layers (RTAL): Re-initialize the weights of the last layer and fine-tune all layers.
\end{itemize}

\textbf{Weight Pruning.} Weight Pruning is usually used to compress a model. It cuts weight connections by setting the weights to zero. We prune $p$ fraction of weights that have the smallest absolute values and then fine-tune the pruned model to recover its accuracy. We increase $p$ from 0.1 to 0.9 with a step size 0.1, a total of 9 pruning rate. For each pruning rate, we create a pruned model. 

\textbf{Weight Noising.} Weight Noising adds noise to the model weights, and fine-tune it to restore model accuracy. It is originally proposed to generate more models with similar weights \cite{metafinger}. We set the weight noising value from 1 to 9 with a step size 1, a total of nine different levels. For each noise level, we create a perturbed model. 

\textbf{Adversarial Training.} Adversarial training augments the training data with adversarial examples to enhance the model robustness, which is the most effective method to defend against adversarial attacks. We implement adversarial training by infusing adversarial examples during fine-tuning. We apply project gradient descent (PGD) \cite{PGD} $L_\infty$ attacks with a norm of 8/255 to generate adversarial examples for $r$ percent of fine-tune data. We set $r$ from 0.1 to 1 with a step size 0.1, a total of ten levels. For each level, we adversarially fine-tune a model. 

\textbf{Distillation.} Distillation is a method originally proposed to distill knowledge from a larger teacher model  to a smaller student model. It introduces a temperature $T$ in the softmax function to scales the teacher DNN’s logit, which increases the prediction’s entropy. Then the scaled logit is used to train the student model. Therefore, the adversary can regard the logit of target model as the scaled logit ($T=1$), then steal a model with distillation. In this setting, the adversary requires accessing to the training dataset. 

\textbf{Knockoff.} Knockoff  trains a substitute model that is functional similar to the target model by only querying the target model. Even when the adversary does not have the knowledge of the trained dataset or model architecture, the adversary can still steal a model.  We implement the Knockoff with the random selection strategy and adopt CIFAR-100 dataset to query the source model. 

Knockoff attack and distillation attack both belong to model extraction, the difference lies in the dataset used to query the model. Distillation attack utilizes the same train dataset as the target model while the knockoff attack uses other different dataset.  


\section{Effectiveness against Knockoff attack}

Fig.~\ref{fig:knockoff} shows the results against knockoff attack. We can observe that all methods achieve better results than distillation attacks. We believe the reason is that knockoff trains the substitute models with another dataset, which reduces the similarity between positive models and negative models.

\bibliographystyle{named}
\bibliography{ijcai23}